\documentclass[10pt,twocolumn,letterpaper]{article}

\usepackage{iccv}
\usepackage{times}
\usepackage{epsfig}
\usepackage{graphicx}
\usepackage{amsmath}
\usepackage{amssymb}
\usepackage{textcomp}

\usepackage{multirow}
\usepackage{makecell}

\usepackage[pagebackref=true,breaklinks=true,letterpaper=true,colorlinks,bookmarks=false]{hyperref}

\iccvfinalcopy 

\def\iccvPaperID{2278} 

\ificcvfinal\fi

\begin{document}

\definecolor{americanrose}{rgb}{1.0, 0.01, 0.24}
\definecolor{green(ryb)}{rgb}{0.4, 0.69, 0.2}
\title{Learning Inner-Group Relations on Point Clouds}

\author{Haoxi Ran\footnotemark[1]~~\footnotemark[2] \quad  Wei Zhuo\footnotemark[2] \quad Jun Liu\footnotemark[2] \quad Li Lu\footnotemark[1] \\
\footnotemark[1]~~Sichuan University \qquad \footnotemark[2]~~Tencent \\ 
{\tt\normalsize \{ranhaoxi, junsenselee\}@gmail.com} \\
{\tt\normalsize weizhuo@tencent.com \quad luli@scu.edu.cn}
}

\maketitle
\ificcvfinal\thispagestyle{empty}\fi

\begin{abstract}

The prevalence of relation networks in computer vision is in stark contrast to underexplored point-based methods. In this paper, we explore the possibilities of local relation operators and survey their feasibility. We propose a scalable and efficient module, called group relation aggregator. The module computes a feature of a group based on the aggregation of the features of the inner-group points weighted by geometric relations and semantic relations. We adopt this module to design our \textbf{RPNet}. We further verify the expandability of RPNet, in terms of both depth and width, on the tasks of classification and segmentation. Surprisingly, empirical results show that wider RPNet fits for classification, while deeper RPNet works better on segmentation. RPNet achieves state-of-the-art for classification and segmentation on challenging benchmarks. We also compare our local aggregator with PointNet++, with around \textbf{30\%} parameters and \textbf{50\%} computation saving. Finally, we conduct experiments to reveal the robustness of RPNet with regard to rigid transformation and noises.

\end{abstract}

\section{Introduction}

Point cloud processing has attracted considerable attention for its advantages in various applications, including autonomous driving, augmented reality, and robotics. Though easily accessible, unlike other visual elements (i.e., images), point clouds can be difficult to learn due to irregularity. 


To duplicate the success of convolutional networks on regular grids \cite{krizhevsky2012imagenet, simonyan2014very}, some prior works change point clouds into multi-view images \cite{gadelha2018multiresolution, feng2018gvcnn} or regular volumes \cite{wu20153d, gadelha2018multiresolution} before convolution. However, image-based projection and voxelization reduce the resolution of point clouds and result in the damage of internal geometric information. These explicit transformations also lead to complex preprocessing and significant computations.  

\begin{figure*}
\begin{center}
\scalebox{0.85}{\includegraphics[width=\textwidth]{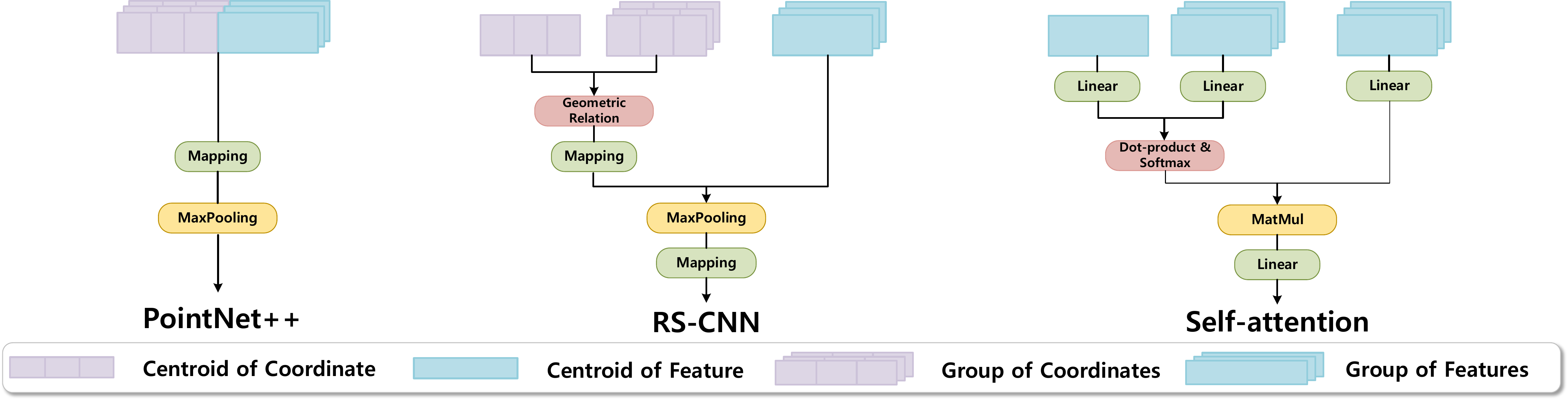}}
\end{center}
   \caption{Comparison of some prior works. We input each module with one centroid (query) and its neighbors (keys) of features or coordinates. PointNet++ \cite{qi2017pointnet++} (left) learns on points independently and lacks the interaction between points. RS-CNN \cite{liu2019relation} (middle) focuses on low-level relation learning, while self-attention (right) in Transformer \cite{vaswani2017attention} learns from high-level relations between features. Here the input group in self-attention is in the global scope.}
\label{fig:compare_priors}
\end{figure*}

PointNet \cite{qi2017pointnet} diverts the attention to the methods of processing raw point clouds. To handle irregular points, it adopts point-wise multi-layer perceptrons (MLP) to learn on points independently and utilize a symmetric function to obtain the global information. For the ignorance of local structures, PointNet++ \cite{qi2017pointnet++} further introduces \textit{set abstraction} (SA) (shown in Fig.~\ref{fig:compare_priors} left) as the local aggregator to build the hierarchical networks. However, this aggregator keeps learning on points independently, losing the sight of \textit{shape awareness}. 

When a local aggregator independently learns on points, the \textit{shape ambiguity} problem has been exposed: since no points inside the set react with others, the aggregator will be sensitive to the coordinates $\mathcal{S} \in \mathbb{R}^{N \times 3}$ and be confused about the outline and the geometric information of the set. Here $N$ is the number of points inside the set. The shape ambiguity problem causes the damage to the robustness and generalization of an aggregator.

In general, an excellent aggregator is underdeveloped for two reasons: it should discriminatively describe the \textit{underlying shape} of point sets, and it should be robust to \textit{rigid transformation} (i.e., translation, rotation) as well as noises.  For a preliminary exploration, RS-CNN \cite{liu2019relation} computes a point feature from the aggregation of features weighted by predefined geometric relations (\textit{low-level relation}) between the point $\mathcal{S}_{i}$ and its neighbors $\mathcal{N}\left(\mathcal{S}_{i}\right)$ (shown in Fig.~\ref{fig:compare_priors} middle). Based on the low-level relations instead of coordinates only, RS-CNN is insensitive to coordinates and robust to rigid transformation. However, RS-CNN is insufficient to learn semantic relations (\textit{high-level relations}) for the lack of interaction between features. 

In this case, self-attention \cite{vaswani2017attention} (shown in Fig.~\ref{fig:compare_priors} right) may be a good instance as the  supplement for high-level relations. Self-attention achieves great success on natural language processing. Recent work \cite{zhao2020exploring,hu2019local} has shown that self-attention can be a viable alternative for convolution on images. However, self-attention can be depressing for \textit{significant computations} as well as \textit{a large number of parameters}. The goal of this work is to extend grid-based self-attention to irregular points with a high-efficiency strategy.  

To this end, we propose \textit{group relation aggregator} (GRA) to learn from both low-level and high-level relations. Compared with self-attention and SA, our designed bottleneck version of GRA is obviously efficient in terms of computation and the number of parameters. With bottleneck GRA, we construct the efficient point-based networks \textbf{RPNet}. 

Specifically, we construct each point set by taking a sampled point $\mathcal{S}_{i}$ as the centroid and its neighbors $\mathcal{N}\left(\mathcal{S}_{i}\right)$, corresponding to features $\mathcal{F}_{i}$ and $\mathcal{N}\left(\mathcal{F}_{i}\right)$ respectively. To exploit a point set, we force GRA to learn attention of the set from both predefined geometric priors (i.e., Euclidean distance between $\mathcal{S}_{i}$ and $\mathcal{N}\left(\mathcal{S}_{i}\right)$) and feature-level interaction (i.e., mapping through a linear layer followed by matrix multiplication and scaled dot-product on $\mathcal{S}_{i}$ and $\mathcal{N}\left(\mathcal{F}_{i}\right)$). By applying the attention to the transformed features $\mathcal{N}\left(\mathcal{F}_{i}\right)$ (i.e.~mapping through a linear layer), the weighted features can reflect the geometric shape as well as semantic information of a point set. This proposed module benefits from the geometric priors in terms of shape awareness and robustness to rigid transformation, while the feature interaction enables the adaptation to content and the robustness to noises. 

Considering the efficiency of GRA, we introduce the bottleneck concept to this module. The output of the first linear layer has the identical channels to the input. We cut down its output channels with a specific factor. Though this behavior harms the model quality (discussed in \cite{vaswani2017attention}), we add a mapping after the relation operation. Cross-channel attention also helps the module to explore channel-wise. All the changes turn our module into a high-efficiency version. 

With our proposed module, we construct RPNet with respect to width (RPNet-W) and depth (RPNet-D). We then evaluate these two types of models on the datasets of classification (i.e., ModelNet40 \cite{wu20153d}) and segmentation (i.e., ScanNet v2 \cite{dai2017scannet}, S3DIS \cite{armeni20163d}). The results show that our method outperforms point-based methods by a large margin, and even achieves comparable performance with all convolution-based methods in a state of high-efficiency. Interestingly, our model may obtain extra accuracy on classification by increasing the width, while deep model works better than wide model on segmentation in terms of efficiency and accuracy.

\begin{figure*}
\begin{center}
\scalebox{0.95}{\includegraphics[width=\textwidth]{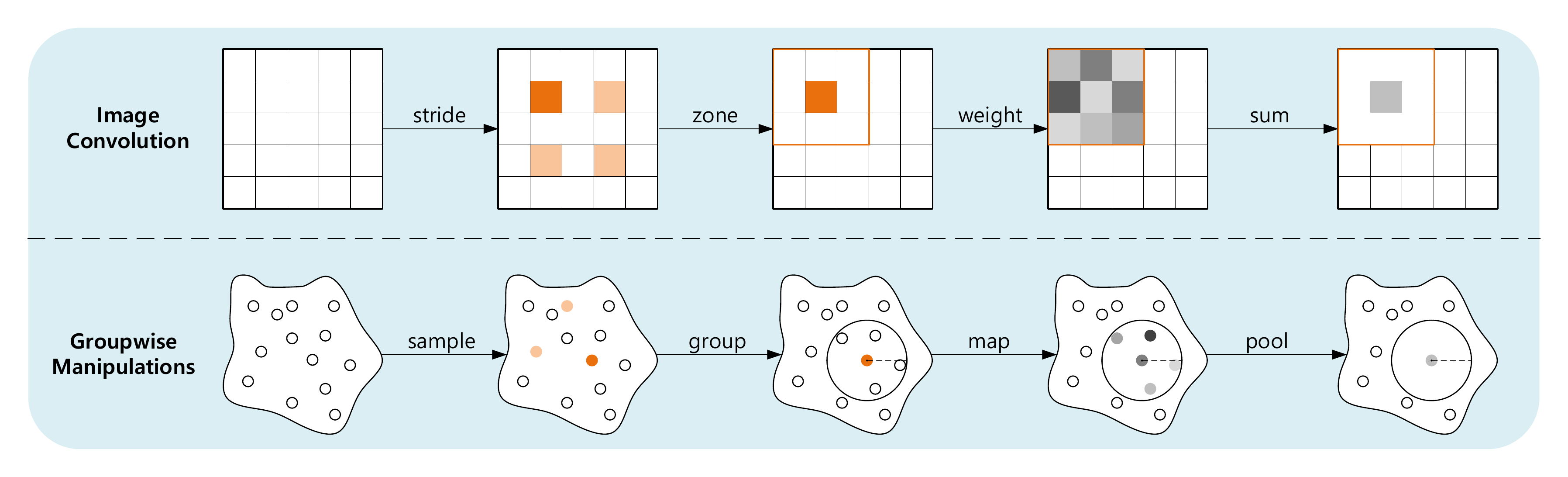}}
\end{center}
   \caption{A simple analogy between image convolution and set abstraction.}
\label{fig:paradigm}
\end{figure*}

Our key contributions are manifold:
\begin{itemize}
\item A novel scalable local aggregator for point clouds is proposed. It encodes the geometric and semantic relations between points;
\item An expandable and high-efficiency hierarchy RPNet is proposed. Equipped with the bottleneck version of our aggregator, extensive RPNet keeps efficient;
\item Experiments on the challenging benchmarks of classification and segmentation, indicate that RPNet achieves state of the arts.
\end{itemize}

\section{Related Work}


\subsection{Learning on Point Clouds}

\textbf{Multi-view methods} \cite{feng2018gvcnn,guo2016multi,xie2016deepshape,han2018seqviews2seqlabels,qi2016volumetric} describe a 3D object with multiple views from different viewpoints. Recent works have been proposed to recognize 3D shapes through convolutional neural networks, i.e., converting 3D shapes to 2D images \cite{su2015multi} or lattice space \cite{su2018splatnet}. However, this transformation results in the loss of shape information for self-occlusion, and a great number of views are required for decent performance. \textbf{Voxel-based methods} \cite{wu20153d,maturana2015voxnet,gadelha2018multiresolution,klokov2017escape,choy20194d, wang2017cnn, riegler2017octnet} apply volumetric CNN to recognize the 3D grids transformed from 3D shapes, i.e., the efficient submanifold sparse convolution \cite{graham20183d}. The input 3D grid is limited to low resolution considering computational cost, leading to the loss of structural information. Different from multi-view and voxel-based methods, the goal of our work is to process point clouds directly.

\textbf{Point-based methods} \cite{lang2020samplenet,fujiwara2020neural,le2020going,liu2020closer,nezhadarya2020adaptive,jiang2019hierarchical} have attracted great attention for processing on raw point clouds recently. PointNet \cite{qi2017pointnet} learns from global information through pointwise multi-layer perceptrons and max-pooling operation. PointNet++ \cite{qi2017pointnet++} introduces set abstraction to capture local features, and farthest point sampling to downsample uniformly between two set abstractions. Likewise, recent works concentrate on effective local learning approaches or various sampling manners. Point2Sequence \cite{liu2019point2sequence} learns the information from different local regions by attention mechanism. ShufflePointNet \cite{chen2020go} goes wider in an effficent manner via group convolution and channel shuffling. RandLA-Net \cite{hu2020randla} aggregates the local region with spatial encoding followed by attentive pooling. \textbf{Convolution-based methods} \cite{xu2018spidercnn,wu2019pointconv,hermosilla2018monte,thomas2019kpconv,liu2019relation,liu2019densepoint,zhang2019shellnet,lin2020convolution,lin2020fpconv,zhao2019pointweb,wang2018deep,wu2019pointconv} are another branch for local aggregation, using dynamic strategies of transformation to support the normal work of convolution on point clouds. PointCNN \cite{li2018pointcnn} applies traditional convolution on point clouds after transforming neighboring points to the canonical order. Grid-GCN \cite{xu2020grid} captures local geometry by graph instead of point set. However, these intuitively predefined transformations for the followed convolution operation may also cause a loss of structural information of the original point clouds, and the model may be sensitive to rigid transformation for convolution. In this paper, we focus on the relation learning based on MLP for a robust learning.

\subsection{Relation Learning}

Relation learning (i.e., self-attention \cite{vaswani2017attention}) has generally revolutionized natural language processing \cite{dai2019transformer,devlin2018bert}. It inspires applications in different computer vision fields, including image recognition \cite{dai2017deformable,wang2017residual,hu2018squeeze,hu2019local,bello2019attention,ramachandran2019stand,yin2020disentangled,zhao2020exploring}, image synthesis \cite{zhang2019self,parmar2018image}, object detection \cite{hu2018relation, gu2018learning}, and video understanding \cite{wang2018non}. Wang~\etal \cite{wang2018non} uses non-local operation to model the relation between two pixels in an image, capturing long-range dependencies. DETR \cite{carion2020end} adopts transformer encoder-decoder architecture for a competitive end-to-end detector.

Recent work proves the practicability of relation learning on point clouds. PointASNL \cite{yan2020pointasnl} adopts the non-local operation to capture long-range dependencies of point clouds. ShapeContextNet \cite{xie2018attentional} applies self-attention-like operator to learn the global feature of a point cloud. Though effective, these methods focus on global relation learning, leading to loss of local information. RS-CNN \cite{liu2019relation} learns the relations within a local region by a predefined geometric priors, but the low-level relation cannot fully represent the relation between two points. In this paper, we aim to design a local aggregator to model the relation between two points on both geometric level and semantic level. 



\section{Relation Learning on Point Clouds}

In this section, we first review PointNet++ \cite{qi2017pointnet++} , and the relation-based modules RS-CNN \cite{liu2019relation} and self-attention \cite{vaswani2017attention}. Next, we propose a general operator to learn inner-group relations and its instances. Then we design its bottleneck version as the building block, and finally, we implement the network architectures, Inner-Group {\bf R}elation {\bf P}oint-based {\bf Net}works (named {\bf RPNet}) with this block.


\subsection{Background}

Most point-based blocks come from SA (shown in Fig.~\ref{fig:compare_priors} left) in PointNet++ \cite{qi2017pointnet++} to aggregate local features. It achieves point downsampling as well as feature transformation. Denote one input point as $x_{i} \in \mathbb{R}^{3 \times N}$, its neighbors as $x_{i \cdot}$, its feature as $\mathbf{f}\left({x_{i}}\right) \in \mathbb{R}^{C \times N}$ and its cooridnate as $\mathbf{p}\left({x_{i}}\right) \in \mathbb{R}^{3 \times N}$, with $C$ being the channels of input point features and $N$ the number of input points. Specifically, this layer transforms the group of feature points $\mathbf{f}\left({x_{i \cdot}}\right)$ via pointwise multi-layer perceptrons followed by max-pooling after point sampling and grouping. For an intuitive presentation, we show an analogy between the operations inside image convolution and SA in Fig.~\ref{fig:paradigm}. SA without sampling can be formulated as follows:
\begin{equation} 
\mathbf{y}\left({x_{i}}\right)=\mathcal{A}\left(\{\mathcal{M}\left(x_{i j}\right),
\forall x_{i j} \in \mathcal{G}\left(x_{i}\right) \}\right),
\end{equation}
where $\mathcal{A}$ is the aggregation function (i.e., max-pooling), $\mathcal{M}$ is the mapping function, and $\mathcal{G}$ is the grouping method (i.e., kNN, Ball Query \cite{qi2017pointnet++}). Here we adopt farthest point sampling as the default sampling method.

\begin{figure}
\begin{center}
\scalebox{0.48}{\includegraphics[width=\textwidth]{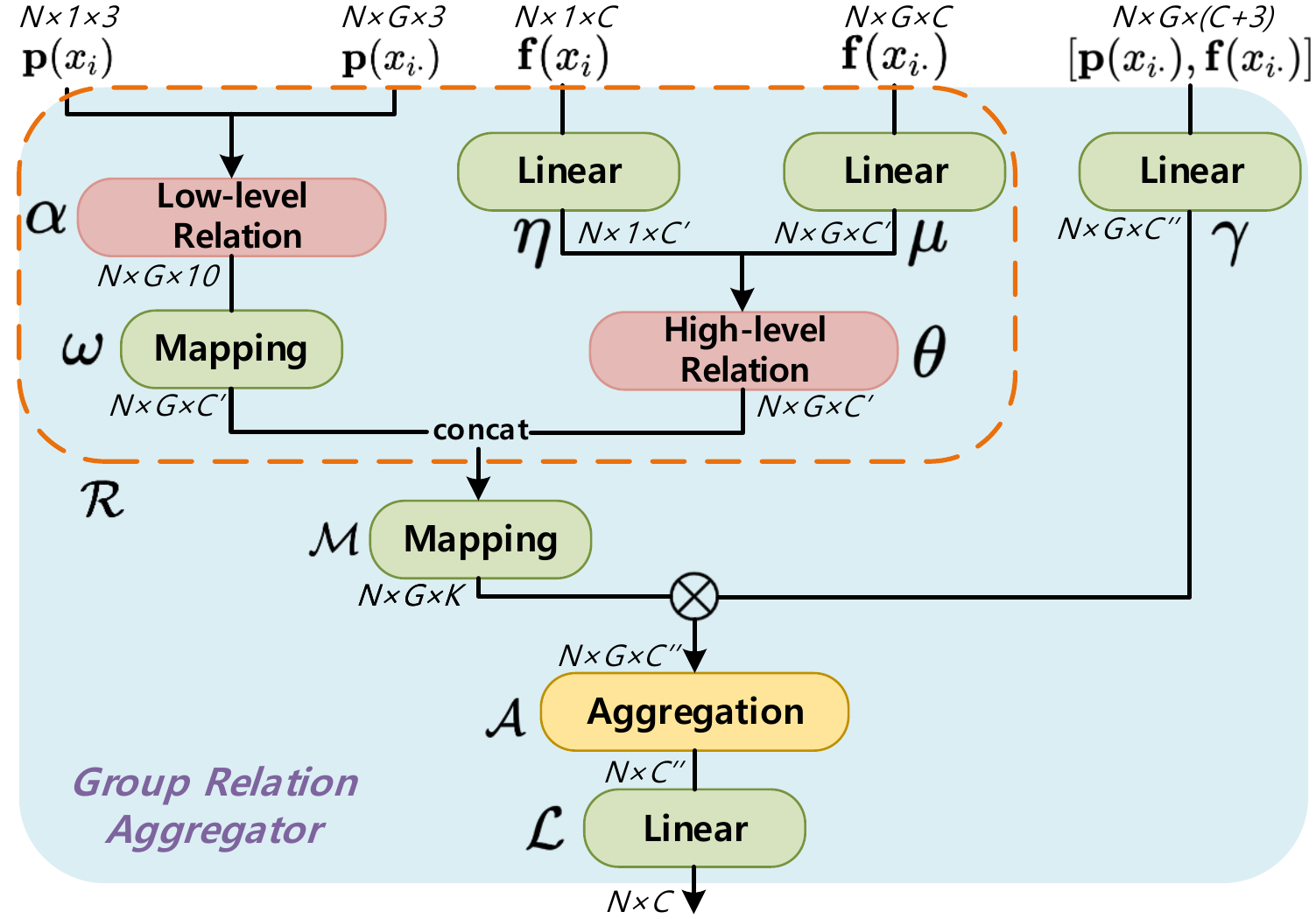}}
\end{center}
   \caption{Our group relation aggregator. After sampling and grouping operations on one point cloud, we input features and coordinates of the centroids $\mathbf{f}\left(x_{i}\right)$, $\mathbf{p}\left(x_{i}\right)$ and their neighbors $\mathbf{f}\left(x_{i \cdot}\right)$, $\mathbf{p}\left(x_{i \cdot}\right)$. There are two parts in the relation function $\mathcal{R}$, $\alpha$ for the geometric relation as well as $\theta$ for the semantic relation. We apply cross-channel attention (denoted as $\otimes$ in the above figure) to the output of $\mathcal{M}$ and the $\gamma$. Here we introduce the cross-channel attention, which segments the output of $\gamma$ to $K$ groups with size $N \times G \times (C'' / K)$, and we apply one of $K$ weight map with size $N \times G$ to each of the $(C'' / K)$ channels of the corresponding group via element-wise product. The outputs of cross-channel attention and $\gamma$ have the same shape. Finally, we aggregate the output followed by a linear function.}
\label{fig:module}
\end{figure}

SA is popular for its simplicity, but it suffers from shape ambiguity as it learns from features or coordinates independently. In this way, RS-CNN \cite{liu2019relation} tries to solve this problem. Relation-shape convolution is of shape-awareness for the predefined geometric relation. As shown in Fig.~\ref{fig:compare_priors} middle, the relation-shape convolution adaptively aggregates the key contents according to the weight from the predefined function $\mathcal{R}_{l}$. For a more powerful shape-aware representation, it further applies a channel-raising mapping $\mathcal{L}_{cr}$ after the weighted features. Relation-shape convolution can be formulated as:
\begin{equation} 
\mathbf{y}\left({x_{i}}\right)=\mathcal{L}_{cr}\left(\mathcal{A}\left(\left\{\mathcal{R}_{l}\left(x_{i}, x_{i j}\right) \cdot \mathbf{f}\left({x_{i j}}\right), \forall x_{i j} \in \mathcal{G}\left(x_{i}\right)\right\}\right)\right).
\end{equation}

Relation-shape convolution is good practice to learn geometric relations on point clouds, but semantic level relations may be ignored. In this case, self-attention operation can be an inspiration to complement the semantic level relation learning on point clouds. Recently, self-attention is proved to be effective in computer vision. Similar to relation-shape convolution, self-attention (shown in Fig.~\ref{fig:compare_priors} right) also learns to aggregate the key contents according to the weight (or compatibility) of query-key pairs. However, differently, the weight is implicitly obtained by the interaction of one query and its keys in a high-level space instead of 3D space. The relation function is defined as $\mathcal{R}_{h}$ (i.e., scaled dot-product). $\mathcal{L}$ is a linear function after aggregation. Self-attention operator can be formulated as:
\begin{equation} 
\mathbf{y}\left({x_{i}}\right)=\mathcal{L}\left(\mathcal{A}\left(\left\{\mathcal{R}_{h}\left(x_{i}, x_{i j}\right) \cdot \beta\left({x_{i j}}\right), \forall x_{i j} \in \mathcal{G}\left(x_{i}\right)\right\}\right)\right).
\end{equation}


\begin{figure*}
\begin{center}
\scalebox{0.85}{\includegraphics[width=\textwidth]{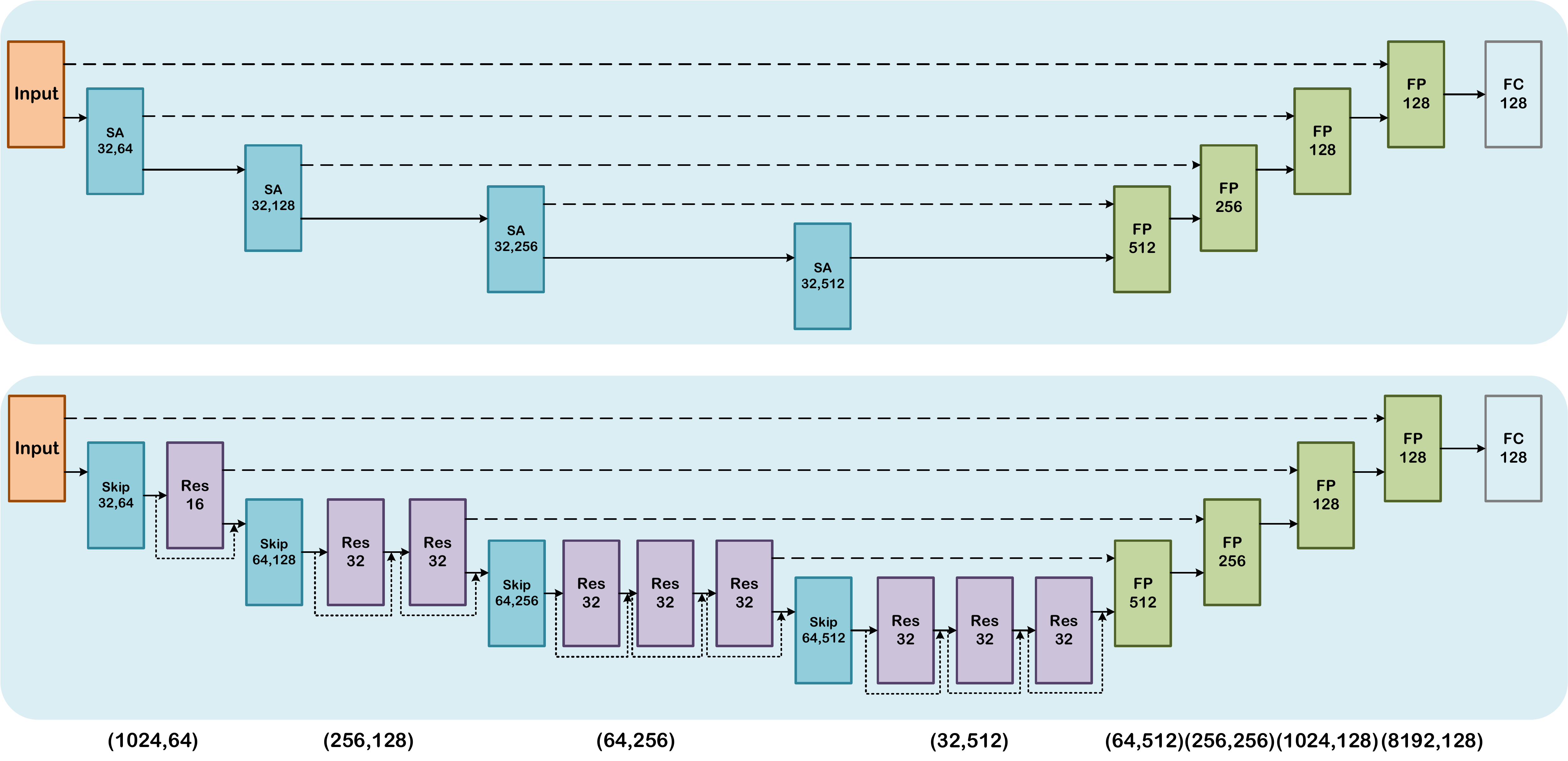}}
\end{center}
   \caption{Comparison of architectures of PointNet++ (above) and our RPNet-D14 (below) for segmentation task. `Skip' and `Res' represent `Skip Block' and `Residual Block' respectively. The two blocks are based on our GRA. Skip block is a GRA combined with down-sampling, while residual block has a branch of residual link. Each skip block groups with a specific scale and outputs the features with a fixed dimension, while each residual block enhances the features through a fixed scale of groups. For example, the first skip block in RPNet-D14 groups 32 points corresponding to the center points and outputs 64 dimension vectors. The following residual block groups 16 points per center point.}
\label{fig:archs}
\end{figure*}

\subsection{Inner-Group Relation Learning}
\label{sec:selfattention}

We explore a local aggregator to learn from both geometric relations and semantic relations inside a point set. Thus we propose group relation aggregator (shown in Fig.~\ref{fig:module}), which has the following form:
\begin{equation} 
\mathbf{y}\left({x_{i}}\right)=\mathcal{L}\left(\mathcal{A}\left(\left\{\mathcal{H}\left(x_{i}, x_{i j}\right), \forall x_{i j} \in \mathcal{G}\left(x_{i}\right)\right\}\right)\right),
\label{eq:1}
\end{equation}
where the inner-group relation function $\mathcal{H}$ equipped with cross-channel attention is defined as:
\begin{equation} 
\begin{split}
\mathcal{H}\left(x_{i}, x_{i j}\right)=\mathcal{M}\left(\mathcal{R}\left(x_{i}, x_{i j}\right)\right) \otimes \gamma\left({x_{i j}}\right),
\end{split}
\end{equation}
where $\otimes$ means cross-channel attention. Denote the number of attention maps as $K$ to enable the operation of cross-channel attention, the total number of centroids as $N$, each centroid with $G$ neighbors, the feature dimension as $C$. $C'=C / r1$ and $C''=C / r2$ (details in Sec.~3.4). We found our introduction of cross-channel attention beneficial by allowing the model to attend to information from different representation subspaces. Cross-channel attention first segments the output of $\gamma$ to $K$ groups with size $N \times G \times (C'' / K)$. For each of $K$ weight map obtained from $\mathcal{M}$, it then applies the map with size $N \times G$ to every channel (totally $C'' / K$ channels) of the corresponding group by element-wise product. The number of channels in the output of $\gamma$ is consistent after the operation of cross-channel attention. Cross-channel attention will degenerate into vanilla attention when $K=1$. The operation of cross-channel attention can further allow the design of bottleneck, significantly reducing the number of parameters and the computations. $\mathcal{M}$ a combination of both linear functions and non-linearity functions, i.e., $\{MLP \rightarrow ReLU \rightarrow MLP \}$. It allows us to introduce additional trainable transformations for more expressive construction of the weights. The output dimensionality of $\omega$ does not need to match that of $\gamma$ as the attention weights are shared across a group of channels for cross-channel attention. The function $\gamma$ is a linear function here. The relation function $\mathcal{R}$ contains a geometric function $\alpha\left(\cdot,\cdot\right)$ followed by a mapping function $\omega$ as well as a semantic function $\theta\left(\cdot,\cdot\right)$:
\begin{equation} 
\mathcal{R}\left(x_{i}, x_{i j}\right)=[\omega\left(\alpha\left(x_{i}, x_{i j}\right)\right), \theta\left(x_{i}, x_{i j}\right)],
\end{equation}
where $\omega$ is a sequence of mapping operations, and $\alpha$ can be defined like this:
\begin{equation} 
\begin{split}
\alpha\left(x_{i}, x_{i j}\right)=[\left\|\mathbf{p}\left(x_{i}\right)-\mathbf{p}\left(x_{i j}\right)\right\|, \mathbf{p}\left(x_{i}\right), \mathbf{p}\left(x_{i j}\right) \\
 \mathbf{p}\left(x_{i}\right) - \mathbf{p}\left(x_{i j}\right)].
\end{split}
\end{equation}

We explore possible instantiations of $\theta$, along with feature transformation elements that surround self-attention operations in our architecture:
\medskip\\
\noindent
{\bf Concatenation:} $\theta\left(x_{i}, x_{i j}\right)=\left[\eta\left(\mathbf{f}\left(x_{i}\right)\right),\mu\left(\mathbf{f}\left(x_{i j}\right)\right)\right]$\medskip\\
{\bf Summation:} $\theta\left(x_{i}, x_{i j}\right)=\eta\left(\mathbf{f}\left({x_{i}}\right)\right)+\mu\left(\mathbf{f}\left(x_{i j}\right)\right)$\medskip\\
{\bf Subtraction:} $\theta\left(x_{i}, x_{i j}\right)=\eta\left(\mathbf{f}\left(x_{i}\right)\right)-\mu\left(\mathbf{f}\left(x_{i j}\right)\right)$\medskip\\
{\bf Hadamard product:} $\theta\left(x_{i}, x_{i j}\right)=\eta\left(\mathbf{f}\left(x_{i}\right)\right) \odot \mu\left(\mathbf{f}\left(x_{i j}\right)\right)$\medskip

Here $\eta$ and $\mu$ are trainable transformations such as linear mappings, and have matching output dimensionality.

\subsection{Bottleneck Improves Efficiency}

Matrix multiplication in self-attention brings about significant computations. The complexity of GRA is of $\mathcal{O}\left(C^{2}\right)$. To design an efficient aggregator, we introduce the philosophy of bottleneck to GRA. Denote the channel dimensionality of input features by $C$. The output of $\eta$ and $\mu$ have $C/r_{1}$ channels. The output of $\gamma$ and $\mathcal{H}$ have the same dimension $C/r_{2}$. The output of the block is subsequently expanded back to $C$ through a linear mapping. In our architectures, we set $r_{1}=16$ and $r_{2}=4$.

\subsection{RPNet}

The main structures of our RPNet generally follow PointNet++ \cite{qi2017pointnet++}, which we use as our baseline. The number $X$ in RPNet-W$X$ and RPNet-D$X$ refers to the number of our GRA blocks.

\textbf{Classification.}~~RPNet-W consists of GRA only. The backbone of RPNet-W has three stages, each with different spatial resolution. Every stage comprises multiple self-attention blocks. In RPNet-W7, grouping of the first two stages are performed by multi-scale groupers, with the sizes of $\{16, 32, 128\}$ and $\{32, 64, 128\}$ in order. RPNet-W9 and RPNet-W15 adopt more detailed scales within $[16, 128]$ and $[32, 128]$. The third stage groups all the rest points for aggregation. The output of the third stage is processed by a classification layer that comprises three linear layers and dropout with a ratio of 0.5 between two of the layers, followed by a softmax activation.

\textbf{Segmentation.}~~We build RPNet-D with skip block (GRA with down-sampling) and residual block (GRA with a residual link). The input of RPNet-D is 8k or 16k points containing various information (i.e.~coordinates, color and normal). RPNet-D first encodes a point cloud by downsampling points, i.e. $\{8192\rightarrow1024\rightarrow256\rightarrow64\rightarrow32\}$, and then decodes by upsampling through feature propagation \cite{qi2017pointnet++}. The residual GRA blocks are attached to the downsampling or upsampling blocks. The segmentation layer includes a linear layer and a dropout layer of 0.5. An illustration (Fig.~\ref{fig:archs}) shows the architecture of RPNet-D14.

\begin{figure}
\begin{center}
\scalebox{0.45}{\includegraphics[width=\textwidth]{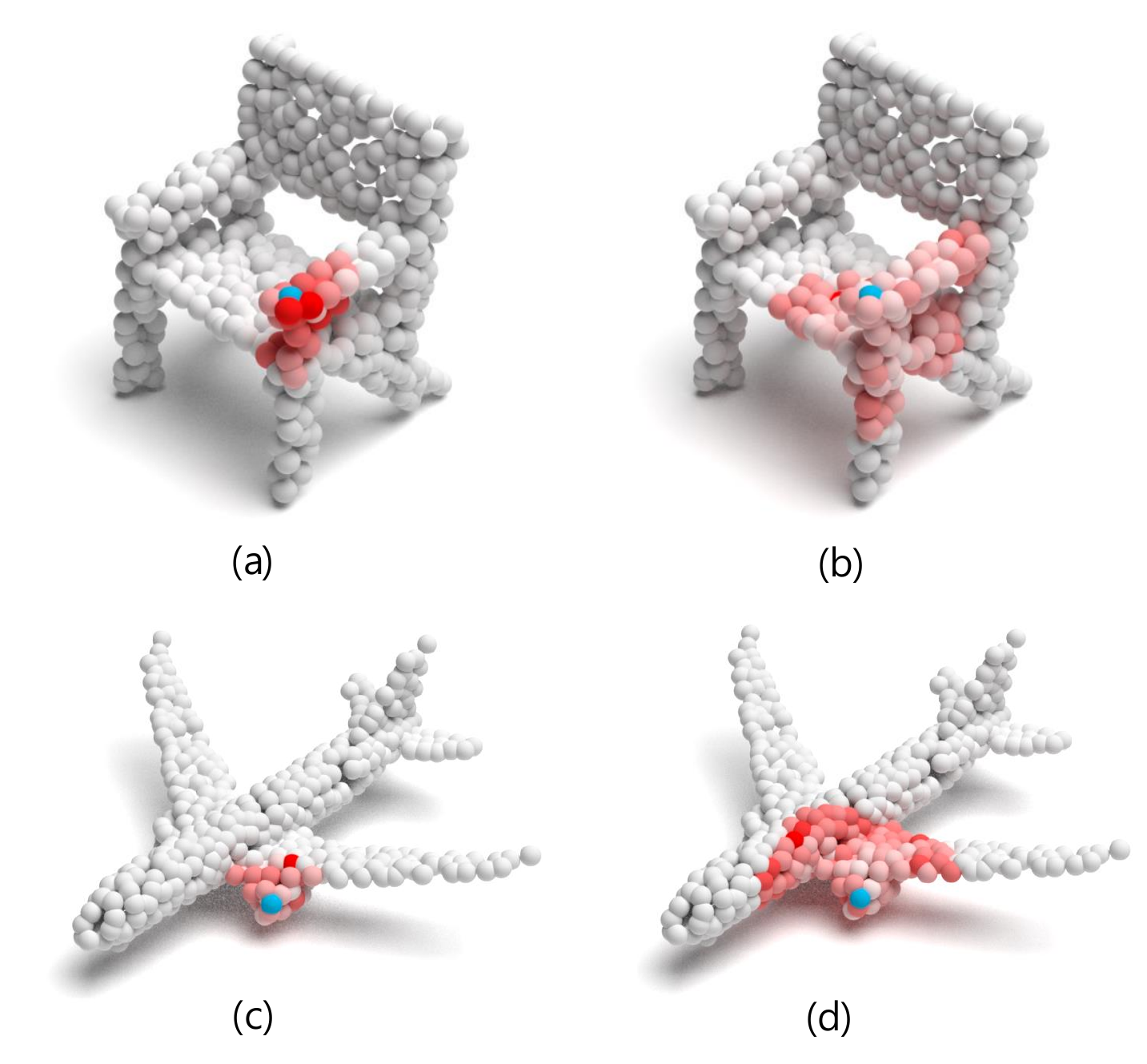}}
\end{center}
   \caption{
We visualize the attention weights of a chair (a)(b) and an airplane (c)(d) from our GRA in different scales. Blue balls stand for the center (query) points. By visualizing, we can find that the bounding points obtain higher importance. We argue that GRA describes a shape representation by the bounding points to a large extent. Intuitively, it works on human eyes as well. We discriminate a shape by its outline most of the time.}
\label{fig:cls}
\end{figure}

\section{Experiments}

First, we evaluate our RPNet-W and RPNet-D on the classification and segmentation tasks of various datasets, including synthetic datasets and scene segmentation datasets. Furthermore, we discuss the expandability of our RPNet. We then analyze the robustness of our models in terms of rigid transformation and noises. The following ablation studies explore the variants of our module and assess the effectiveness of the components used to construct the networks. All the experiments are performed on a machine with four V100 GPUs.

\subsection{Evaluation on Classification}

We conduct experiments on ModelNet40 \cite{wu20153d}  on classification through our RPNet-W network. The dataset contains 9843 training point clouds and 2468 test ones from 40 different categories.

\textbf{Implementation.}~~Our implementation mainly follows the practice in \cite{qi2017pointnet}. For training, we first select 1024 points as input. To prevent overfitting, we apply augmentation strategy including the following components: random scaling in the range $[0.8, 1.25]$, random shift in the range $[-0.1, 0.1]$, random dropout points with the ratio of the range $[0, 87.5\%]$. The initial learning rate is 0.001, and it decays by a factor of 0.7 every 20 epochs. For testing, similar to \cite{qi2017pointnet,qi2017pointnet++}, we average the predictions of randomly scaled inputs.

\textbf{Results.}~~In Tab.~\ref{tab:modelnet}, we compare our RPNet-W with state-of-the-art classification methods on ModelNet40. Among all current methods, our method achieves state-of-the-art with a promotion of 0.9\%. We also compare our RPNet-W7 with RS-CNN and PointASNL with and without normal input. Besides, we visualize the attention maps in Fig.~\ref{fig:cls}.

\begin{table}[]
\begin{center}
\scalebox{0.85}{
\begin{tabular}{l|c|cl}
\hline
Method      & Modality         & Accuracy(\%) \\ \hline
PointGCN \cite{zhang2018graph}    & Graph         & 89.5     \\
KPConv \cite{thomas2019kpconv} & Grid      & 92.9     \\
SO-Net \cite{li2018so}   & Points+Normals (5k) & 93.4     \\ 
PVRNet  \cite{you2019pvrnet}    & Points+Views   & 93.6     \\ 
RS-CNN \cite{liu2019relation} w/ vot.   & Points        & 93.6     \\ 
\hline\hline
PointNet \cite{qi2017pointnet}   & Points        & 89.2     \\
RS-CNN \cite{liu2019relation} w/o vot.   & Points        & 92.4     \\ 
PointASNL \cite{yan2020pointasnl}  & Points        & 92.9     \\
\hline
RPNet-W7 (\textbf{ours})  & Points        & ~~~~~~~\textbf{93.8} \textcolor{green(ryb)}{\small $\uparrow$0.9}    \\ 
RPNet-W9 (\textbf{ours})  & Points &  ~~~~~~~\textbf{93.9} \textcolor{green(ryb)}{\small $\uparrow$1.0}    \\ 
\hline\hline
PointNet++ \cite{qi2017pointnet++}  & Points+Normals & 91.9     \\ 
FPConv \cite{lin2020fpconv} & Points+Normals       & 92.5     \\
Grid-GCN \cite{xu2020grid}   & Points+Normals         & 93.1     \\
PointASNL \cite{yan2020pointasnl}  & Points+Normals & 93.2     \\ 
\hline
RPNet-W7 (\textbf{ours})   & Points+Normals &   ~~~~~~~\textbf{93.9} \textcolor{green(ryb)}{\small $\uparrow$0.7}   \\ 
RPNet-W9 (\textbf{ours}) & Points+Normals & ~~~~~~~\textbf{94.1} \textcolor{green(ryb)}{\small $\uparrow$0.9}    \\ 
\hline
\end{tabular}}
\end{center}
\caption{Performance of classification on ModelNet40 on accuracy(\%).}
\label{tab:modelnet}
\end{table}

\begin{table}[]
\begin{center}
\scalebox{0.85}{
\begin{tabular}{l|c|c}
\hline
Method & S3DIS-6 & ScanNet \\ 
\hline\hline
\multicolumn{3}{c}{\textit{Convolution-based Methods}} \\
\hline
PointCNN     \cite{li2018pointcnn}           & 65.4              & 45.8                     \\
FPConv     \cite{lin2020fpconv}             & 68.7          & 63.9                     \\
KPConv  \cite{thomas2019kpconv}   & \textbf{70.6}         & \textbf{68.4}            \\ 
\hline\hline
\multicolumn{3}{c}{\textit{MLP-based Methods}} \\
\hline
RandLA       \cite{hu2020randla} ($10^{5}$)          & 70.0               & -                        \\
PointNet++     \cite{qi2017pointnet++}         & 53.4           & 33.9                     \\
PointWeb     \cite{zhao2019pointweb}           & 66.7               & -                        \\
PointASNL   \cite{yan2020pointasnl}             & 68.7            & 63.0                     \\
\hline
RPNet-D8  (\textbf{ours})            &     69.1            &     67.1                    \\
RPNet-D14  (\textbf{ours})            &     70.0            &     67.7                  \\
RPNet-D27  (\textbf{ours})   &      ~~~~~~~\textbf{70.8} \textcolor{green(ryb)}{\small $\uparrow$2.1}              &      ~~~~~~~\textbf{68.2} \textcolor{green(ryb)}{\small $\uparrow$5.2}  \\ 
\hline
\end{tabular}}
\end{center}
\caption{Mean per-class IoU(\%) for the task of semantic segmentation on the datasets of ScanNet~v2 and S3DIS (6-fold cross validation). ``-" means unknown.}
\label{tab:semantic}
\end{table}

\subsection{Evaluation on Segmentation}

Large-scale scene segmentation is a more challenging task due to outliers and noises. We evaluate our RPNet-D on Stanford 3D Large-Scale Indoor Spaces (\textit{S3DIS}) \cite{armeni20163d} and ScanNet v2 (\textit{ScanNet}) \cite{dai2017scannet} datasets. \textit{S3DIS} contains 271 scenes from six zones. It provides 13 types of semantic labels for scene segmentation. \textit{ScanNet} includes 1513 training point clouds and 100 test ones. It marks each point from 21 categories.

\textbf{Implementation.}~~On both datasets, we verify each method with mean per-class IoU (mIoU), and use point position and RGB information as input. In particular, we evaluate models with 6-fold cross-validation over all six zones (6-fold) on \textit{S3DIS}. For training, we randomly sample 16384 points from the scenes. For evaluation, similar to \cite{yan2020pointasnl}, we obtain an average prediction of 5 votes by sliding a window across the room in 0.5m stride.

\begin{figure}
\begin{center}
\scalebox{0.45}{\includegraphics[width=\textwidth]{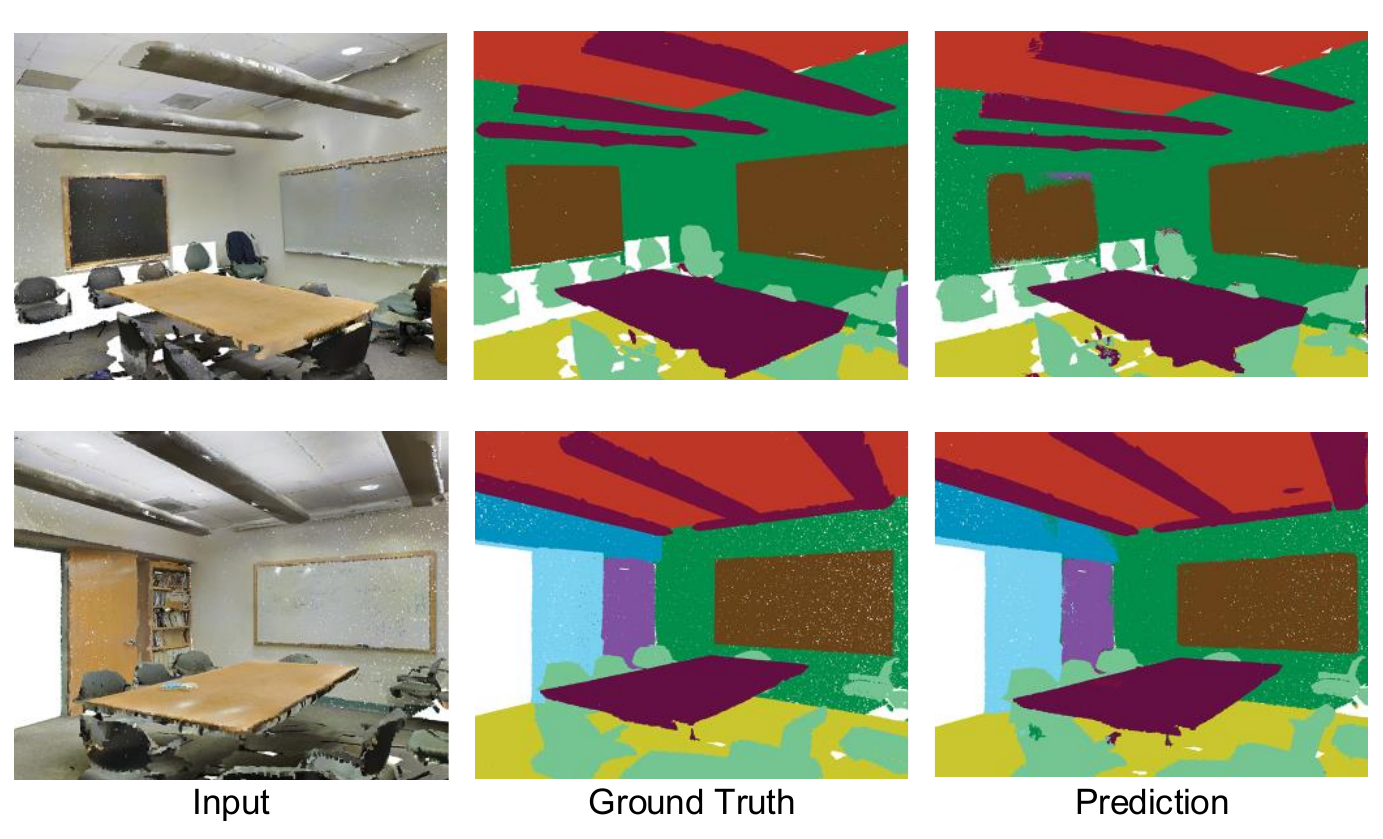}}
\end{center}
   \caption{Examples of semantic scene labeling with RPNet.}
\label{fig:vis_semantic}
\end{figure}

\textbf{Results.}~~In Tab.~\ref{tab:semantic}, we list the latest methods to compare with our RPNet-D, i.e., PointASNL \cite{yan2020pointasnl}. We adopt the same training and test approach by randomly chopping cubes with a fixed number of points. We show the results of convolution-based methods as well, i.e., FPconv \cite{lin2020fpconv}, KPconv \cite{thomas2019kpconv}. All of these methods utilize the raw point clouds as the input. An illustration of semantic scene labeling is shown in Fig.~\ref{fig:vis_semantic}.

\subsection{Discussion of Expandability}
\label{sec:wider_deeper}

\begin{figure}
\begin{center}
\scalebox{0.5}{\includegraphics[width=\textwidth]{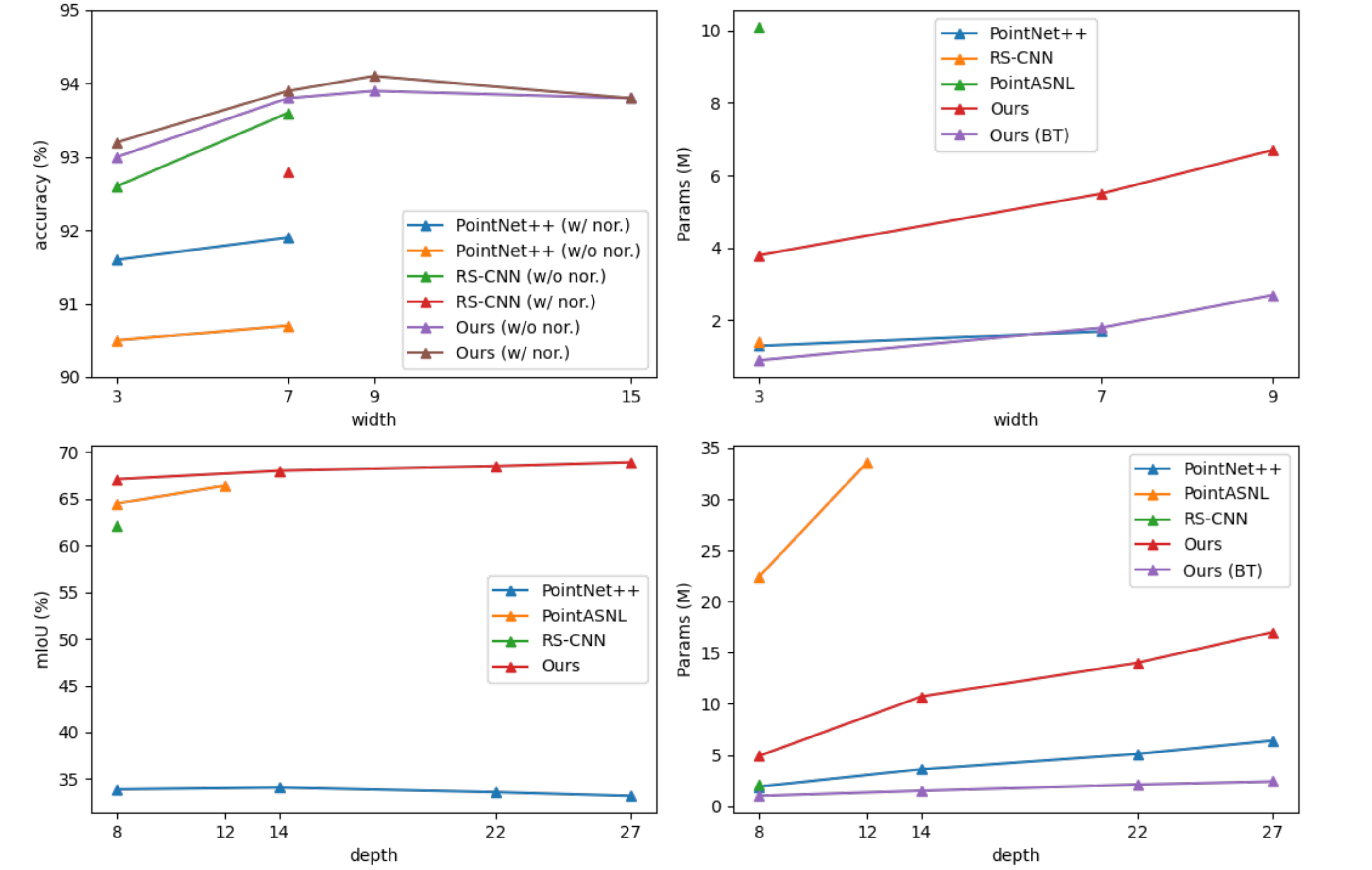}}
\end{center}
   \caption{Model comparison in terms of width on classification (above) and depth on segmentation (below). We compare models by both accuracy or mIoU (left) and the number of parameters (right). Here ``nor.'' means normals, ``BT'' means bottleneck version.}
\label{fig:plot_both}
\end{figure}

As show in Fig.~\ref{fig:plot_both}, we evaluate various models with respect to accuracy and the number of parameters. With the same depth or width, our model outperforms SOTAs by an obvious margin. Besides, the efficiency of our model is also competitive compared with prior works.

Shown in Fig.~\ref{fig:plot_both} above, we compare our RPNet-W with PointNet++ multi-scale grouping (MSG) version. To improve efficiency, we use a bottleneck architecture to build blocks. These blocks significantly reduce the computational cost while maintaining high accuracy. By increasing the width, our model can obtain more geometric information from a point cloud, and thus higher accuracy. The comparison of RPNet-W9 and RPNet-W15 shows a overfitting problem. We argue that the width of 9 is enough for the full exploitation. Wider model would bring much redundant information and cause the decrease of performance. 

Shown in Fig.~\ref{fig:plot_both} below, similar to \cite{le2020going}, we deepen our models on segmentation. We compare the efficiency improvement of RPNet-D with our bottleneck blocks. To verify the effectiveness of greater depth, we gradually increase the depth of RPNet-D on ScanNet. Obviously, as the depth increases, the model can obtain higher mIoU. We argue that relation learning helps to deeply extract representations by stacking residual blocks. We also obtain a similar conclusion from \cite{le2020going} that going deeper can help improve the accuracy on segmentation.

To explore which kind of model works better on the two tasks, we conduct simple experiments on deeper models for classification and wider models for segmentation. First, we test deeper model using RPNet-W1 on classification:
\\[4pt]
\centerline{
\begin{tabular}{c|ccc}
Depth & 3  & 5 & 7 \\ 
\Xhline{3\arrayrulewidth}
acc. & 92.9 & 92.5 \textcolor{americanrose}{\small $\downarrow$0.4} & 92.1 \textcolor{americanrose}{\small $\downarrow$0.8}            
\end{tabular}} 
\\[4pt]
Shown in the table, the results perform worse with the depth increasing. We discuss that the task of classification concentrates on the global view instead of point-wise recognition. Increasing the depth could not be beneficial to obtain such global view. Besides, there would be a great number of redundant features inside each layer. The errors of a global view would increase significantly if the depth is over the threshold value. 

Also, we test wider RPNet-D4 (RPNet-D without residual block) on segmentation:
\\[4pt]
\centerline{
\begin{tabular}{l|cccc}
Width & 1 & 2 & 3 \\ 
\Xhline{3\arrayrulewidth}
acc. & 66.8 & 67.0 \textcolor{green(ryb)}{\small $\uparrow$0.2} & 67.1 \textcolor{green(ryb)}{\small $\uparrow$0.3}           
\end{tabular}} 
\\[4pt]
Shown in the table, increasing the width leads to almost no improvement, but with nearly $X$-fold increasing on the number of parameters and the computation. Through the empirical results as well as our discussion, we conclude that RPNet-W fits for classification, while RPNet-D works better on segmentation.

\subsection{Ablation Study}

We conduct ablation studies to evaluate the effectiveness of our design. We mainly discuss some key components of our GRA, including inner-group relation function, aggregation function and cross-channel attention. We perform all ablation studies on S3DIS with 6-fold validation.

\begin{table}[]
\begin{center}
\scalebox{0.7}{
\begin{tabular}{c|cccc|cccc|c}
\hline
\multirow{2}{*}{Model} & \multicolumn{4}{c|}{Geometric Relation $\alpha$}                        & \multicolumn{4}{c|}{Semantic Relation $\theta$}        & \multirow{2}{*}{mIoU (\%)} \\ \cline{2-9}
                       & $\ell 2$ & $\ell 1$ & $x_{i}-x_{ij}$ & \multicolumn{1}{c|}{$[x_{i}, x_{ij}]$} & sum & sub & cat & \multicolumn{1}{c|}{had}  &                       \\ \hline
A                      & \checkmark  &    &        &                                    &     &     &     &                          &        63.7               \\
B                      & \checkmark  & \checkmark  &        &                                    &     &     &     &                          &      63.8               \\
C                      & \checkmark  & \checkmark  & \checkmark      & \checkmark                                  &     &     &     &                          &          64.2           \\
D                      & \checkmark  & \checkmark  & \checkmark      & \checkmark                                  & \checkmark   &     &     &                          &     \textbf{67.8}                    \\
E                      & \checkmark  & \checkmark  & \checkmark      & \checkmark                                  &     & \checkmark   &     &                          &  67.8                     \\
F                      & \checkmark  & \checkmark  & \checkmark      & \checkmark                                  &     &     & \checkmark   &                          &     67.5                 \\
G                      & \checkmark  & \checkmark  & \checkmark      & \checkmark                                  &     &     &     & \checkmark                        &        67.6               \\
\hline
\end{tabular}}
\end{center}
\caption{The results of different designs on inner-group relation function $\mathcal{H}$ (sum: summation, sub: subtraction, cat: concatenation, had: Hadamard product). The experiments are on S3DIS with 6-fold validation.}
\label{tab:ablation_relation}
\end{table}

\textbf{Inner-group relation function $\mathcal{H}$.}~
Shown in Tab.~\ref{tab:ablation_relation}, we ablate the designs of geometric and semantic relation functions $\alpha$ and $\theta$ in details. We define the geometric priors as four possible components: $\ell 2$, $\ell 1$, $x_{i}-x_{ij}$ and $[x_{i}, x_{ij}]$. $\ell 2$ and $\ell 1$ can directly describe the distances between two points, while $x_{i}-x_{ij}$ and $[x_{i}, x_{ij}]$ show the relative and global positions of two points. Model C outperforms model A and B, which shows that four components of geometric priors boost the performance of our RPNet at the same time. Furthermore, we survey the designs of semantic relation function $\theta$ with four possibilities: summation, subtraction, concatenation and Hadamard product. The results prove that summation or subtraction works better in RPNet. Summation or subtraction would be better in terms of computation as well. 

\textbf{Aggregation function $\mathcal{A}$.}~~
We adopt three types of symmetric function to aggregate in our GRA: max-pooling, avg-pooling and sum-pooling. Here is the results:
\\[4pt]
\centerline{
\begin{tabular}{c|ccc}
RPNet-D8 & max-pooling  & avg-pooling & sum-pooling \\ 
\Xhline{3\arrayrulewidth}
acc. & \textbf{67.8}  & 67.5 &   67.7       
\end{tabular}} 
\\[4pt]
The table shows that max-pooling performs better than the others. We argue that max-pooling can filter the redundant information in our GRA and select the expressive features.

\textbf{Cross-channel attention.}~~
We test the design of our cross-channel attention below:
\\[4pt]
\centerline{
\begin{tabular}{c|cc}
RPNet-D8 & vanilla att.  & cross-channel att.\\ 
\Xhline{3\arrayrulewidth}
acc. & 67.8 & \textbf{69.1} \textcolor{green(ryb)}{\small $\uparrow$1.3}        
\end{tabular}} 
\\[4pt]
As shown in the table, cross-channel attention greatly boosts our GRA by 1.3\% on S3DIS. Vanilla attention does not enable the channelwise exploration. However, different channels may play different roles to influence the final weights. One channel attention map cannot fully utilize the feature space, especially when the space has a great number of dimensions. To handle this problem, cross-channel attention applies multiple attention maps to different groups of channels, allowing the channelwise exploration in GRA.

\begin{table}[]
\begin{center}
\scalebox{0.85}{
\begin{tabular}{l|c|ccccc}
\hline
model           & origin & perm & $\pm0.2$ & 90\textdegree   & 180\textdegree  & 270\textdegree  \\ \hline
PointNet \cite{qi2017pointnet}       & 88.7   & 88.7 & 70.8 & 42.5 & 38.6 & 40.7 \\
PointNet++ \cite{qi2017pointnet++}     & 88.2   & 88.2 & 88.2 & 88.2 & 47.9 & 39.7 \\
RS-CNN  \cite{liu2019relation}        & 90.3   & 90.3 & 90.3 & 90.3 & 90.3 & 90.3 \\ 
\textbf{RPNet-W7 (ours)} & \textbf{90.9}   &\textbf{90.9} & \textbf{90.9} & \textbf{90.9} & \textbf{90.9}& \textbf{90.9} \\ \hline
\end{tabular}}
\end{center}
\caption{Robustness to point permutation and rigid transformation (\%). We perform the operations of random permutation (perm), translation of $\pm0.2$, and clockwise rotation around Y axis.}
\label{tab:robust_rt}
\end{table}

\subsection{Analysis of Robustness}

\begin{figure}
\begin{center}
\scalebox{0.4}{\includegraphics[width=\textwidth]{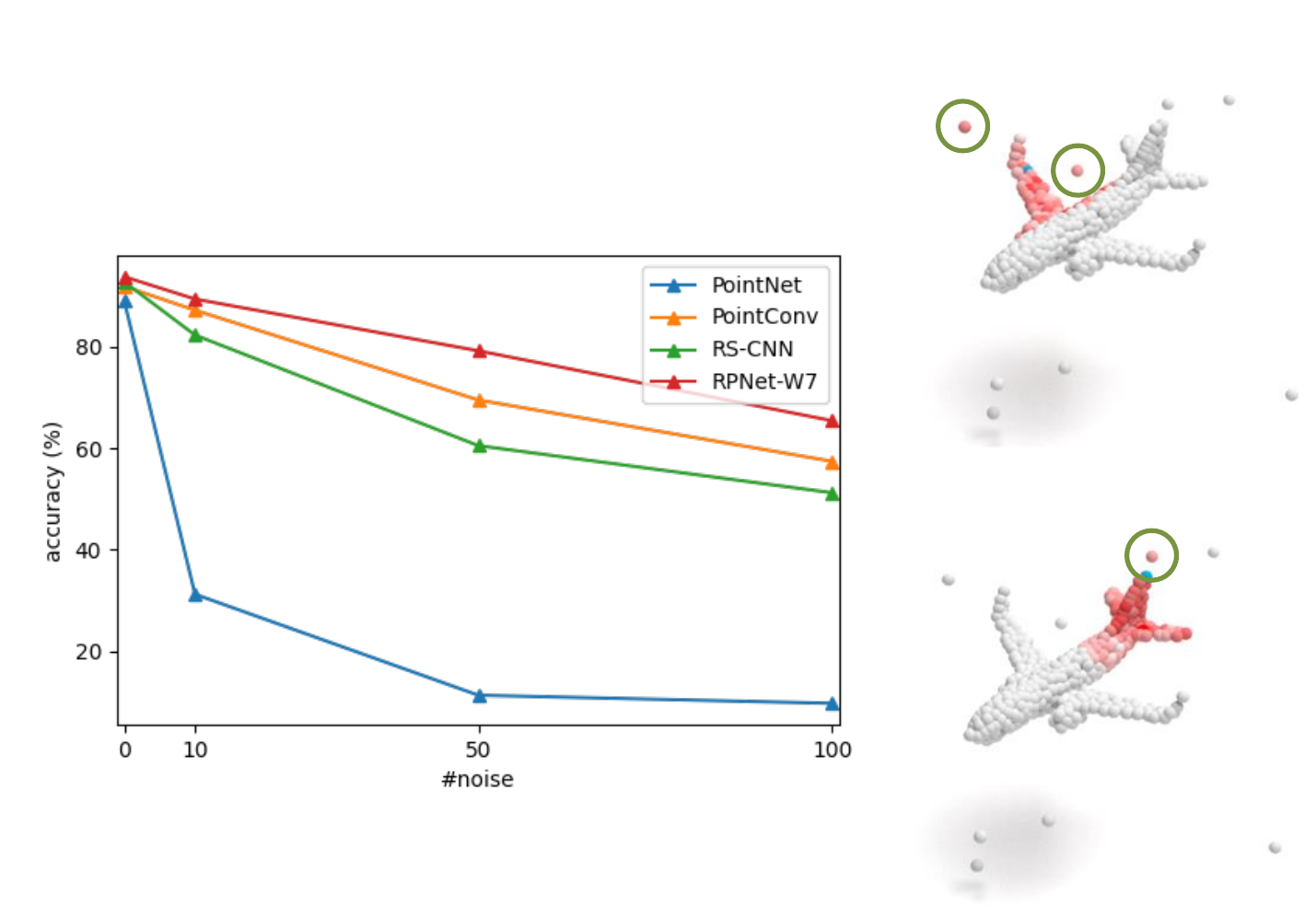}}
\end{center}
   \caption{(Left) Classification results of different models with noises. (Right) By learning semantic relation, the local aggregator is relatively insensitive to noises, concentrating more on the shape instead of independent points.}
\label{fig:plot_noise}
\end{figure}

\textbf{Robustness to rigid transformation.}~
We evaluate the robustness to rigid transformation for the comparison of our RPNet with PointNet \cite{qi2017pointnet}, PointNet++ \cite{qi2017pointnet++}, RS-CNN \cite{liu2019relation}. We follow the experimental setup of RS-CNN for the evaluation. As shown in Tab.~\ref{tab:robust_rt}, all these coordinate-based methods are insensitive to permutation thanks to the design of symmetric function. However, PointNet is sensitive to translation and rotation, while PointNet++ is vulnerable to rotation. RS-CNN and our RPNet perform robust due to the usage of relation learning. Our RPNet outperforms RS-CNN for the learning on high-level relation.

\textbf{Robustness to noises.}~~
We also evaluate our networks on robustness to noises. Shown in Fig.~\ref{fig:plot_noise}, our RPNet outperforms other competitive methods with the same noises input. Note that RS-CNN is sensitive to noise. We argue that the predefined relation may affect the output of attention map, causing the wrong relation representation. However, our RPNet uses semantic level relation, which is skilled at denoising and focusing on the content. 

\section{Conclusion}
We present group relation aggregator as well as deeper (RPNet-D) and wider (RPNet-W) models for efficient point cloud analysis. By learning from both geometric and semantic relations inside a point set, our RPNet achieves state-of-the-art on both classification and segmentation tasks. We further introduce bottleneck philosophy to our module for high efficiency. Experiments based on challenging benchmarks illustrate the effectiveness of our RPNet.

{\small
\bibliographystyle{ieee_fullname}
\bibliography{egbib}

\begin{thebibliography}{10}\itemsep=-1pt

\bibitem{armeni20163d}
Iro Armeni, Ozan Sener, Amir~R Zamir, Helen Jiang, Ioannis Brilakis, Martin
  Fischer, and Silvio Savarese.
\newblock 3d semantic parsing of large-scale indoor spaces.
\newblock In {\em Proceedings of the IEEE Conference on Computer Vision and
  Pattern Recognition}, pages 1534--1543, 2016.

\bibitem{bello2019attention}
Irwan Bello, Barret Zoph, Ashish Vaswani, Jonathon Shlens, and Quoc~V Le.
\newblock Attention augmented convolutional networks.
\newblock In {\em Proceedings of the IEEE International Conference on Computer
  Vision}, pages 3286--3295, 2019.

\bibitem{carion2020end}
Nicolas Carion, Francisco Massa, Gabriel Synnaeve, Nicolas Usunier, Alexander
  Kirillov, and Sergey Zagoruyko.
\newblock End-to-end object detection with transformers.
\newblock {\em arXiv preprint arXiv:2005.12872}, 2020.

\bibitem{chen2020go}
Can Chen, Luca Zanotti~Fragonara, and Antonios Tsourdos.
\newblock Go wider: An efficient neural network for point cloud analysis via
  group convolutions.
\newblock {\em Applied Sciences}, 10(7):2391, 2020.

\bibitem{choy20194d}
Christopher Choy, JunYoung Gwak, and Silvio Savarese.
\newblock 4d spatio-temporal convnets: Minkowski convolutional neural networks.
\newblock In {\em Proceedings of the IEEE/CVF Conference on Computer Vision and
  Pattern Recognition}, pages 3075--3084, 2019.

\bibitem{dai2017scannet}
Angela Dai, Angel~X Chang, Manolis Savva, Maciej Halber, Thomas Funkhouser, and
  Matthias Nie{\ss}ner.
\newblock Scannet: Richly-annotated 3d reconstructions of indoor scenes.
\newblock In {\em Proceedings of the IEEE Conference on Computer Vision and
  Pattern Recognition}, pages 5828--5839, 2017.

\bibitem{dai2017deformable}
Jifeng Dai, Haozhi Qi, Yuwen Xiong, Yi Li, Guodong Zhang, Han Hu, and Yichen
  Wei.
\newblock Deformable convolutional networks.
\newblock In {\em Proceedings of the IEEE international conference on computer
  vision}, pages 764--773, 2017.

\bibitem{dai2019transformer}
Zihang Dai, Zhilin Yang, Yiming Yang, Jaime Carbonell, Quoc~V Le, and Ruslan
  Salakhutdinov.
\newblock Transformer-xl: Attentive language models beyond a fixed-length
  context.
\newblock {\em arXiv preprint arXiv:1901.02860}, 2019.

\bibitem{devlin2018bert}
Jacob Devlin, Ming-Wei Chang, Kenton Lee, and Kristina Toutanova.
\newblock Bert: Pre-training of deep bidirectional transformers for language
  understanding.
\newblock {\em arXiv preprint arXiv:1810.04805}, 2018.

\bibitem{feng2018gvcnn}
Yifan Feng, Zizhao Zhang, Xibin Zhao, Rongrong Ji, and Yue Gao.
\newblock Gvcnn: Group-view convolutional neural networks for 3d shape
  recognition.
\newblock In {\em Proceedings of the IEEE Conference on Computer Vision and
  Pattern Recognition}, pages 264--272, 2018.

\bibitem{fujiwara2020neural}
Kent Fujiwara and Taiichi Hashimoto.
\newblock Neural implicit embedding for point cloud analysis.
\newblock In {\em Proceedings of the IEEE/CVF Conference on Computer Vision and
  Pattern Recognition}, pages 11734--11743, 2020.

\bibitem{gadelha2018multiresolution}
Matheus Gadelha, Rui Wang, and Subhransu Maji.
\newblock Multiresolution tree networks for 3d point cloud processing.
\newblock In {\em Proceedings of the European Conference on Computer Vision
  (ECCV)}, pages 103--118, 2018.

\bibitem{graham20183d}
Benjamin Graham, Martin Engelcke, and Laurens Van Der~Maaten.
\newblock 3d semantic segmentation with submanifold sparse convolutional
  networks.
\newblock In {\em Proceedings of the IEEE conference on computer vision and
  pattern recognition}, pages 9224--9232, 2018.

\bibitem{gu2018learning}
Jiayuan Gu, Han Hu, Liwei Wang, Yichen Wei, and Jifeng Dai.
\newblock Learning region features for object detection.
\newblock In {\em Proceedings of the European Conference on Computer Vision
  (ECCV)}, pages 381--395, 2018.

\bibitem{guo2016multi}
Haiyun Guo, Jinqiao Wang, Yue Gao, Jianqiang Li, and Hanqing Lu.
\newblock Multi-view 3d object retrieval with deep embedding network.
\newblock {\em IEEE Transactions on Image Processing}, 25(12):5526--5537, 2016.

\bibitem{han2018seqviews2seqlabels}
Zhizhong Han, Mingyang Shang, Zhenbao Liu, Chi-Man Vong, Yu-Shen Liu, Matthias
  Zwicker, Junwei Han, and CL~Philip Chen.
\newblock Seqviews2seqlabels: Learning 3d global features via aggregating
  sequential views by rnn with attention.
\newblock {\em IEEE Transactions on Image Processing}, 28(2):658--672, 2018.

\bibitem{hermosilla2018monte}
Pedro Hermosilla, Tobias Ritschel, Pere-Pau V{\'a}zquez, {\`A}lvar Vinacua, and
  Timo Ropinski.
\newblock Monte carlo convolution for learning on non-uniformly sampled point
  clouds.
\newblock {\em ACM Transactions on Graphics (TOG)}, 37(6):1--12, 2018.

\bibitem{hu2018relation}
Han Hu, Jiayuan Gu, Zheng Zhang, Jifeng Dai, and Yichen Wei.
\newblock Relation networks for object detection.
\newblock In {\em Proceedings of the IEEE Conference on Computer Vision and
  Pattern Recognition}, pages 3588--3597, 2018.

\bibitem{hu2019local}
Han Hu, Zheng Zhang, Zhenda Xie, and Stephen Lin.
\newblock Local relation networks for image recognition.
\newblock In {\em Proceedings of the IEEE International Conference on Computer
  Vision}, pages 3464--3473, 2019.

\bibitem{hu2018squeeze}
Jie Hu, Li Shen, and Gang Sun.
\newblock Squeeze-and-excitation networks.
\newblock In {\em Proceedings of the IEEE conference on computer vision and
  pattern recognition}, pages 7132--7141, 2018.

\bibitem{hu2020randla}
Qingyong Hu, Bo Yang, Linhai Xie, Stefano Rosa, Yulan Guo, Zhihua Wang, Niki
  Trigoni, and Andrew Markham.
\newblock Randla-net: Efficient semantic segmentation of large-scale point
  clouds.
\newblock In {\em Proceedings of the IEEE/CVF Conference on Computer Vision and
  Pattern Recognition}, pages 11108--11117, 2020.

\bibitem{jiang2019hierarchical}
Li Jiang, Hengshuang Zhao, Shu Liu, Xiaoyong Shen, Chi-Wing Fu, and Jiaya Jia.
\newblock Hierarchical point-edge interaction network for point cloud semantic
  segmentation.
\newblock In {\em Proceedings of the IEEE International Conference on Computer
  Vision}, pages 10433--10441, 2019.

\bibitem{klokov2017escape}
Roman Klokov and Victor Lempitsky.
\newblock Escape from cells: Deep kd-networks for the recognition of 3d point
  cloud models.
\newblock In {\em Proceedings of the IEEE International Conference on Computer
  Vision}, pages 863--872, 2017.

\bibitem{krizhevsky2012imagenet}
Alex Krizhevsky, Ilya Sutskever, and Geoffrey~E Hinton.
\newblock Imagenet classification with deep convolutional neural networks.
\newblock {\em Advances in neural information processing systems},
  25:1097--1105, 2012.

\bibitem{lang2020samplenet}
Itai Lang, Asaf Manor, and Shai Avidan.
\newblock Samplenet: Differentiable point cloud sampling.
\newblock In {\em Proceedings of the IEEE/CVF Conference on Computer Vision and
  Pattern Recognition}, pages 7578--7588, 2020.

\bibitem{le2020going}
Eric-Tuan Le, Iasonas Kokkinos, and Niloy~J Mitra.
\newblock Going deeper with lean point networks.
\newblock In {\em Proceedings of the IEEE/CVF Conference on Computer Vision and
  Pattern Recognition}, pages 9503--9512, 2020.

\bibitem{li2018so}
Jiaxin Li, Ben~M Chen, and Gim Hee~Lee.
\newblock So-net: Self-organizing network for point cloud analysis.
\newblock In {\em Proceedings of the IEEE conference on computer vision and
  pattern recognition}, pages 9397--9406, 2018.

\bibitem{li2018pointcnn}
Yangyan Li, Rui Bu, Mingchao Sun, Wei Wu, Xinhan Di, and Baoquan Chen.
\newblock Pointcnn: Convolution on x-transformed points.
\newblock In {\em Advances in neural information processing systems}, pages
  820--830, 2018.

\bibitem{lin2020fpconv}
Yiqun Lin, Zizheng Yan, Haibin Huang, Dong Du, Ligang Liu, Shuguang Cui, and
  Xiaoguang Han.
\newblock Fpconv: Learning local flattening for point convolution.
\newblock In {\em Proceedings of the IEEE/CVF Conference on Computer Vision and
  Pattern Recognition}, pages 4293--4302, 2020.

\bibitem{lin2020convolution}
Zhi-Hao Lin, Sheng-Yu Huang, and Yu-Chiang~Frank Wang.
\newblock Convolution in the cloud: Learning deformable kernels in 3d graph
  convolution networks for point cloud analysis.
\newblock In {\em Proceedings of the IEEE/CVF Conference on Computer Vision and
  Pattern Recognition}, pages 1800--1809, 2020.

\bibitem{liu2019point2sequence}
Xinhai Liu, Zhizhong Han, Yu-Shen Liu, and Matthias Zwicker.
\newblock Point2sequence: Learning the shape representation of 3d point clouds
  with an attention-based sequence to sequence network.
\newblock In {\em Proceedings of the AAAI Conference on Artificial
  Intelligence}, volume~33, pages 8778--8785, 2019.

\bibitem{liu2019densepoint}
Yongcheng Liu, Bin Fan, Gaofeng Meng, Jiwen Lu, Shiming Xiang, and Chunhong
  Pan.
\newblock Densepoint: Learning densely contextual representation for efficient
  point cloud processing.
\newblock In {\em Proceedings of the IEEE/CVF International Conference on
  Computer Vision}, pages 5239--5248, 2019.

\bibitem{liu2019relation}
Yongcheng Liu, Bin Fan, Shiming Xiang, and Chunhong Pan.
\newblock Relation-shape convolutional neural network for point cloud analysis.
\newblock In {\em Proceedings of the IEEE Conference on Computer Vision and
  Pattern Recognition}, pages 8895--8904, 2019.

\bibitem{liu2020closer}
Ze Liu, Han Hu, Yue Cao, Zheng Zhang, and Xin Tong.
\newblock A closer look at local aggregation operators in point cloud analysis.
\newblock In {\em European Conference on Computer Vision}, pages 326--342.
  Springer, 2020.

\bibitem{maturana2015voxnet}
Daniel Maturana and Sebastian Scherer.
\newblock Voxnet: A 3d convolutional neural network for real-time object
  recognition.
\newblock In {\em 2015 IEEE/RSJ International Conference on Intelligent Robots
  and Systems (IROS)}, pages 922--928. IEEE, 2015.

\bibitem{nezhadarya2020adaptive}
Ehsan Nezhadarya, Ehsan Taghavi, Ryan Razani, Bingbing Liu, and Jun Luo.
\newblock Adaptive hierarchical down-sampling for point cloud classification.
\newblock In {\em Proceedings of the IEEE/CVF Conference on Computer Vision and
  Pattern Recognition}, pages 12956--12964, 2020.

\bibitem{parmar2018image}
Niki Parmar, Ashish Vaswani, Jakob Uszkoreit, {\L}ukasz Kaiser, Noam Shazeer,
  Alexander Ku, and Dustin Tran.
\newblock Image transformer.
\newblock {\em arXiv preprint arXiv:1802.05751}, 2018.

\bibitem{qi2017pointnet}
Charles~R Qi, Hao Su, Kaichun Mo, and Leonidas~J Guibas.
\newblock Pointnet: Deep learning on point sets for 3d classification and
  segmentation.
\newblock In {\em Proceedings of the IEEE conference on computer vision and
  pattern recognition}, pages 652--660, 2017.

\bibitem{qi2016volumetric}
Charles~R Qi, Hao Su, Matthias Nie{\ss}ner, Angela Dai, Mengyuan Yan, and
  Leonidas~J Guibas.
\newblock Volumetric and multi-view cnns for object classification on 3d data.
\newblock In {\em Proceedings of the IEEE conference on computer vision and
  pattern recognition}, pages 5648--5656, 2016.

\bibitem{qi2017pointnet++}
Charles~Ruizhongtai Qi, Li Yi, Hao Su, and Leonidas~J Guibas.
\newblock Pointnet++: Deep hierarchical feature learning on point sets in a
  metric space.
\newblock In {\em Advances in neural information processing systems}, pages
  5099--5108, 2017.

\bibitem{ramachandran2019stand}
Prajit Ramachandran, Niki Parmar, Ashish Vaswani, Irwan Bello, Anselm Levskaya,
  and Jonathon Shlens.
\newblock Stand-alone self-attention in vision models.
\newblock {\em arXiv preprint arXiv:1906.05909}, 2019.

\bibitem{riegler2017octnet}
Gernot Riegler, Ali Osman~Ulusoy, and Andreas Geiger.
\newblock Octnet: Learning deep 3d representations at high resolutions.
\newblock In {\em Proceedings of the IEEE conference on computer vision and
  pattern recognition}, pages 3577--3586, 2017.

\bibitem{simonyan2014very}
Karen Simonyan and Andrew Zisserman.
\newblock Very deep convolutional networks for large-scale image recognition.
\newblock {\em arXiv preprint arXiv:1409.1556}, 2014.

\bibitem{su2018splatnet}
Hang Su, Varun Jampani, Deqing Sun, Subhransu Maji, Evangelos Kalogerakis,
  Ming-Hsuan Yang, and Jan Kautz.
\newblock Splatnet: Sparse lattice networks for point cloud processing.
\newblock In {\em Proceedings of the IEEE conference on computer vision and
  pattern recognition}, pages 2530--2539, 2018.

\bibitem{su2015multi}
Hang Su, Subhransu Maji, Evangelos Kalogerakis, and Erik Learned-Miller.
\newblock Multi-view convolutional neural networks for 3d shape recognition.
\newblock In {\em Proceedings of the IEEE international conference on computer
  vision}, pages 945--953, 2015.

\bibitem{thomas2019kpconv}
Hugues Thomas, Charles~R Qi, Jean-Emmanuel Deschaud, Beatriz Marcotegui,
  Fran{\c{c}}ois Goulette, and Leonidas~J Guibas.
\newblock Kpconv: Flexible and deformable convolution for point clouds.
\newblock In {\em Proceedings of the IEEE International Conference on Computer
  Vision}, pages 6411--6420, 2019.

\bibitem{vaswani2017attention}
Ashish Vaswani, Noam Shazeer, Niki Parmar, Jakob Uszkoreit, Llion Jones,
  Aidan~N Gomez, {\L}ukasz Kaiser, and Illia Polosukhin.
\newblock Attention is all you need.
\newblock In {\em Advances in neural information processing systems}, pages
  5998--6008, 2017.

\bibitem{wang2017residual}
Fei Wang, Mengqing Jiang, Chen Qian, Shuo Yang, Cheng Li, Honggang Zhang,
  Xiaogang Wang, and Xiaoou Tang.
\newblock Residual attention network for image classification.
\newblock In {\em Proceedings of the IEEE conference on computer vision and
  pattern recognition}, pages 3156--3164, 2017.

\bibitem{wang2017cnn}
Peng-Shuai Wang, Yang Liu, Yu-Xiao Guo, Chun-Yu Sun, and Xin Tong.
\newblock O-cnn: Octree-based convolutional neural networks for 3d shape
  analysis.
\newblock {\em ACM Transactions on Graphics (TOG)}, 36(4):1--11, 2017.

\bibitem{wang2018deep}
Shenlong Wang, Simon Suo, Wei-Chiu Ma, Andrei Pokrovsky, and Raquel Urtasun.
\newblock Deep parametric continuous convolutional neural networks.
\newblock In {\em Proceedings of the IEEE Conference on Computer Vision and
  Pattern Recognition}, pages 2589--2597, 2018.

\bibitem{wang2018non}
Xiaolong Wang, Ross Girshick, Abhinav Gupta, and Kaiming He.
\newblock Non-local neural networks.
\newblock In {\em Proceedings of the IEEE conference on computer vision and
  pattern recognition}, pages 7794--7803, 2018.

\bibitem{wu2019pointconv}
Wenxuan Wu, Zhongang Qi, and Li Fuxin.
\newblock Pointconv: Deep convolutional networks on 3d point clouds.
\newblock In {\em Proceedings of the IEEE Conference on Computer Vision and
  Pattern Recognition}, pages 9621--9630, 2019.

\bibitem{wu20153d}
Zhirong Wu, Shuran Song, Aditya Khosla, Fisher Yu, Linguang Zhang, Xiaoou Tang,
  and Jianxiong Xiao.
\newblock 3d shapenets: A deep representation for volumetric shapes.
\newblock In {\em Proceedings of the IEEE conference on computer vision and
  pattern recognition}, pages 1912--1920, 2015.

\bibitem{xie2016deepshape}
Jin Xie, Guoxian Dai, Fan Zhu, Edward~K Wong, and Yi Fang.
\newblock Deepshape: Deep-learned shape descriptor for 3d shape retrieval.
\newblock {\em IEEE transactions on pattern analysis and machine intelligence},
  39(7):1335--1345, 2016.

\bibitem{xie2018attentional}
Saining Xie, Sainan Liu, Zeyu Chen, and Zhuowen Tu.
\newblock Attentional shapecontextnet for point cloud recognition.
\newblock In {\em Proceedings of the IEEE Conference on Computer Vision and
  Pattern Recognition}, pages 4606--4615, 2018.

\bibitem{xu2020grid}
Qiangeng Xu, Xudong Sun, Cho-Ying Wu, Panqu Wang, and Ulrich Neumann.
\newblock Grid-gcn for fast and scalable point cloud learning.
\newblock In {\em Proceedings of the IEEE/CVF Conference on Computer Vision and
  Pattern Recognition}, pages 5661--5670, 2020.

\bibitem{xu2018spidercnn}
Yifan Xu, Tianqi Fan, Mingye Xu, Long Zeng, and Yu Qiao.
\newblock Spidercnn: Deep learning on point sets with parameterized
  convolutional filters.
\newblock In {\em Proceedings of the European Conference on Computer Vision
  (ECCV)}, pages 87--102, 2018.

\bibitem{yan2020pointasnl}
Xu Yan, Chaoda Zheng, Zhen Li, Sheng Wang, and Shuguang Cui.
\newblock Pointasnl: Robust point clouds processing using nonlocal neural
  networks with adaptive sampling.
\newblock In {\em Proceedings of the IEEE/CVF Conference on Computer Vision and
  Pattern Recognition}, pages 5589--5598, 2020.

\bibitem{yin2020disentangled}
Minghao Yin, Zhuliang Yao, Yue Cao, Xiu Li, Zheng Zhang, Stephen Lin, and Han
  Hu.
\newblock Disentangled non-local neural networks.
\newblock {\em arXiv preprint arXiv:2006.06668}, 2020.

\bibitem{you2019pvrnet}
Haoxuan You, Yifan Feng, Xibin Zhao, Changqing Zou, Rongrong Ji, and Yue Gao.
\newblock Pvrnet: Point-view relation neural network for 3d shape recognition.
\newblock In {\em Proceedings of the AAAI Conference on Artificial
  Intelligence}, volume~33, pages 9119--9126, 2019.

\bibitem{zhang2019self}
Han Zhang, Ian Goodfellow, Dimitris Metaxas, and Augustus Odena.
\newblock Self-attention generative adversarial networks.
\newblock In {\em International Conference on Machine Learning}, pages
  7354--7363. PMLR, 2019.

\bibitem{zhang2018graph}
Yingxue Zhang and Michael Rabbat.
\newblock A graph-cnn for 3d point cloud classification.
\newblock In {\em 2018 IEEE International Conference on Acoustics, Speech and
  Signal Processing (ICASSP)}, pages 6279--6283. IEEE, 2018.

\bibitem{zhang2019shellnet}
Zhiyuan Zhang, Binh-Son Hua, and Sai-Kit Yeung.
\newblock Shellnet: Efficient point cloud convolutional neural networks using
  concentric shells statistics.
\newblock In {\em Proceedings of the IEEE/CVF International Conference on
  Computer Vision}, pages 1607--1616, 2019.

\bibitem{zhao2020exploring}
Hengshuang Zhao, Jiaya Jia, and Vladlen Koltun.
\newblock Exploring self-attention for image recognition.
\newblock In {\em Proceedings of the IEEE/CVF Conference on Computer Vision and
  Pattern Recognition}, pages 10076--10085, 2020.

\bibitem{zhao2019pointweb}
Hengshuang Zhao, Li Jiang, Chi-Wing Fu, and Jiaya Jia.
\newblock Pointweb: Enhancing local neighborhood features for point cloud
  processing.
\newblock In {\em Proceedings of the IEEE Conference on Computer Vision and
  Pattern Recognition}, pages 5565--5573, 2019.

\end{thebibliography}
}

\end{document}